# SAD: A Large-scale Dataset towards Airport Detection in Synthetic Aperture Radar Images

Daochang Wang, Fan Zhang, Fei Ma, Wei Hu, Yu Tang, and Yongsheng Zhou

*Abstract*—Airports have an important role in both military and civilian domains. The synthetic aperture radar (SAR) based airport detection has received increasing attention in recent years. However, due to the high cost of SAR imaging and annotation process, there is no publicly available SAR dataset for airport detection. As a result, deep learning methods have not been fully used in airport detection tasks. To provide a benchmark for airport detection research in SAR images, this paper introduces a large-scale SAR Airport Dataset (SAD). In order to adequately reflect the demands of real world applications, it contains 624 SAR images from Sentinel 1B and covers 104 airfield instances with different scales, orientations and shapes. The experiments of multiple deep learning approach on this dataset proves its effectiveness. It developing state-of-the-art airport area detection algorithms or other relevant tasks.

## I. Introduction

AIRPORTS are high valuable civilian and military facilities where aircraft take off, land, and park. The detection of airports plays an important role in many practical applications. For example, aircraft usually park at the airports, so accurate airport detection can make aircraft detection much easier. Synthetic aperture radar (SAR) has the characteristics of all-weather and all-day imaging on the Earth's surface. At present, airport detection in SAR images is already receiving extensive concern.

There are four major types of methods for airport detection in SAR images, including line-based, image segmentation-based, saliency-based, and deep learning-based. The difference in backscatter intensity between the airport runway and the surrounding environment results in significant line segments around the airport runway. The line-based methods utilize these line segments to detect the airport areas. Line segments are mainly detected by Hough transform [1], Radon transform, or line segment detector (LSD) [2]. After obtaining the line segments, the airport area detection is achieved by judging the length of the line segments [3], [4], [5]. There are also some improved methods. They obtain the airport support regions by line segments and then select the right regions as the airports based on some prior information [6], [7]. Salience-based methods aim at predicting a saliency map, where the airport regions will have high saliency values. The saliency maps are commonly calculated based on line segments [8], [9], [10], [18], [20], frequency domain features [20] and superpixel-wise geometric information [21], [22], [23]. These two types of methods generally have high requirements for line segment detection methods. However, it is difficult to distinguish runways from rivers, roads, and strip lakes, especially under the interference of multiplicative noise in SAR images. The segmentation-based approaches identify the airport regions according to the texture features [12], [13], [14], [15], structural features [11] and intensity information [16]. Nevertheless, experiments show that these methods usually have a heavy computational burden.

In recent years, with the development of deep learning theories [17], deep learning-based object detection methods have been gradually applied to airport detection. Existing airport detection methods use deep learning-based object detection models as the framework. Then some optimizations, such as changing the anchor scale or employing image segmentation modules, are performed on this basis to achieve more accurate localization [32], [33]. These detection models [34] can be divided into two main categories: two-stage and single-stage. The two-stage detection models mainly include RCNN [35], Fast RCNN [36], Faster RCNN [37], and Mask RCNN [38]. Previous airport detection models are mainly based on the two-stage models. For example, Yin et al. [29] used Faster RCNN with the addition of hard sample mining to solve the problem of sample imbalance due to insufficient datasets. Chen et al. [30] also toke Faster RCNN as the basic detection network and modified the scales of convolution kernels and the anchor sizes to achieve more accurate airport area localization. Zeng et al. [31] added a separate CNN after Faster RCNN to further remove the false alarms. The main single-stage detection models include Yolo V3 [39], Yolo V4 [40], and Yolo V5. Up to now, few methods use single-stage detection models to detect airport areas. The main reason is that single-stage object detection models localize and classify the object only once, which makes it difficult to achieve satisfactory detection results with small training data sets. The deep learning-based airport detection methods all require large scale datasets [27]. Since so far, due to the high cost of SAR imaging and difficult annotating process, there is no publicly available SAR dataset for airport detection. The validation experiments of the above methods are on their own SAR airport datasets, which makes it difficult to compare their performance. In addition, these datasets have bias issues, i.e. they tend to use images in ideal conditions (clear backgrounds and without densely distributed instances), which cannot adequately reflect the demands of real world applications. Some approaches first pre-train their models using natural images or optical aerial images. [24], [26]. Other methods address the sample imbalance of data by hard sample mining [28], [29]. However, the difference between SAR images and natural images makes transfer learning unsatisfactory. Applying hard sample mining on a small dataset also does not help the model to fit adequately.

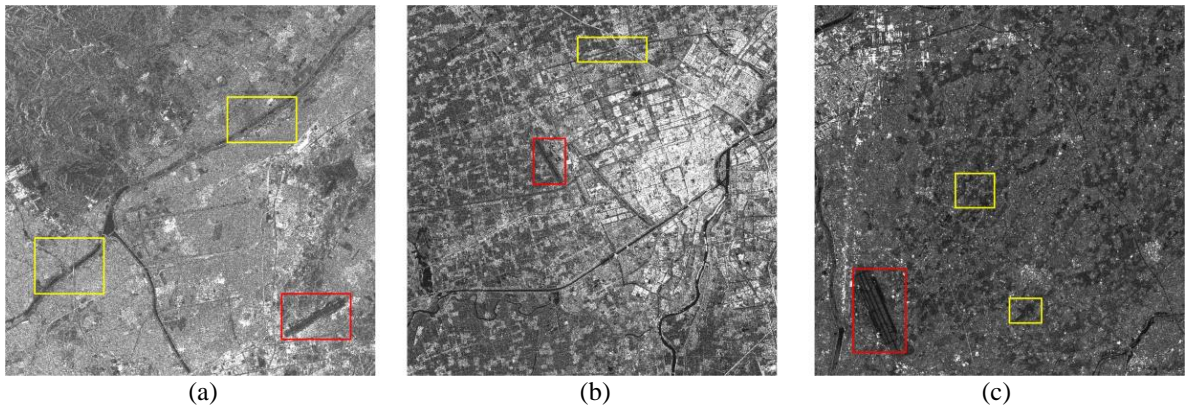

**Fig. 1.** Demonstration of airport areas and airport analogues in SAR images (red rectangles: airports areas, yellows: airport analogues). (a) River areas. (b) Road areas. (c) Strip lakes.

To advance the airport detection research in SAR images, this paper introduces a large-scale SAR Airport Dataset (SAD). We collect 624 SAR images from Sentinel 1B. Each image is of the size about 2048×2048 pixels and contains airports of different scales, orientations and shapes. These SAD images are annotated by experts in aerial image interpretation and contain 104 airfield instances, each of which is labeled by an axis-aligned bounding box.

The main contributions of this article are summarized as follows.

1) To our knowledge, SAD is the first publicly available SAR airport dataset. It provides a benchmark resource for developing state-of-the-art airport area detection algorithms or other relevant tasks.

## II. SAD: SAR AIRPORT DATASET

In this section, we will present the details of the SAR airport dataset.

### A. Images Collection of SAD

Images in SAD are collected from Sentinel 1B. Sentinel 1B is equipped with a C-band SAR sensor. It can provide measurement data for land, forest, ocean, and glacier monitoring and mapping. In this study, we use Level-1 Interferometric Wide Swath GRD Products. The spatial resolution is 10 m × 10 m and the incidence angle is about 20° ~45°. The polarization modes are VV and VH.

To increase the diversity of data, we collect images shot in multiple airports carefully chosen by experts in aerial image interpretation. There are 624 airfield images of 2048 × 2048 pixels in size in SAD, which cover 104 airfield targets. For the images of the same airport, their imaging time is different. Among all airfield targets, 91 airports are located in China and their geographical distributions are shown in Fig. 3. To be specific, they are respectively located in Anhui, Beijing, Fujian, Guangzhou, Hebei, Henan, Heilongjiang, Jilin, Jiangsu, Jiangxi, Shandong, Shanghai, Sichuan, Tianjin, Zhejiang, and Chongqing province. The remaining 13 airports are located in Russia, Korea, India, Mongolia, Australia, United States, Japan, and Indonesia.

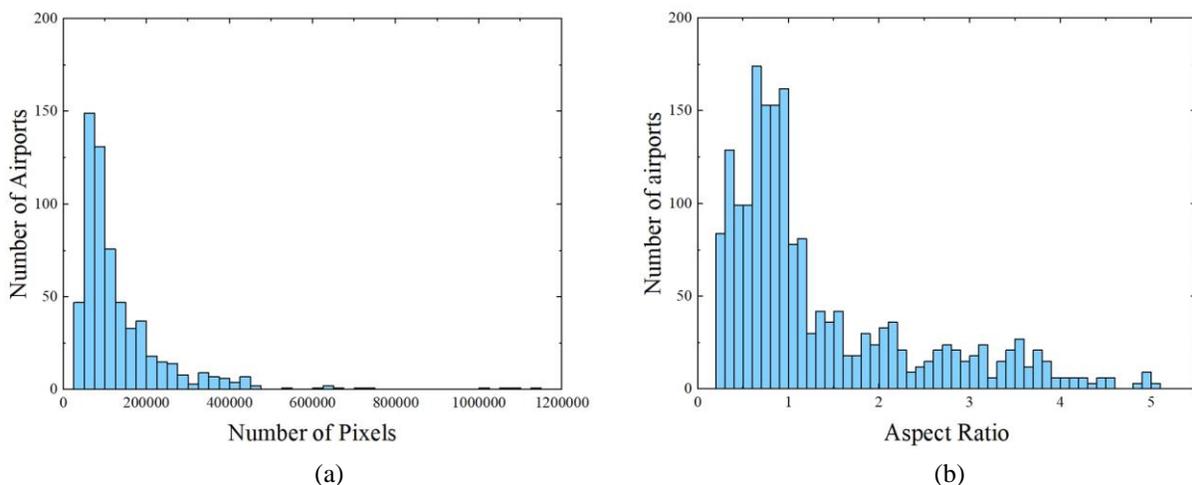

**Fig. 2.** Histogram statistics of the airport area in SAD. (a) Histogram of the number of pixels. (b) Histogram of aspect ratio.

12

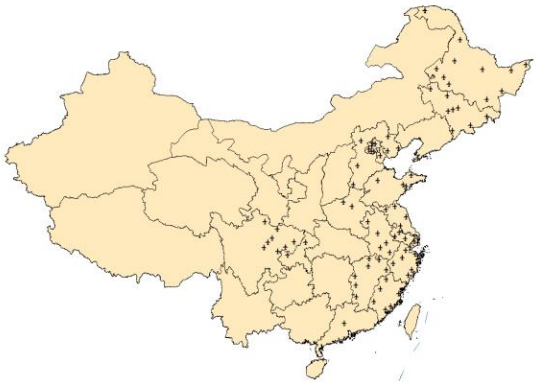

**Fig. 3.** Distribution of airports in China.

### B. Annotation Method

Considering that the distribution of airport targets is not dense, we use horizontal bounding boxes for annotation. The usual representation of horizontal bounding boxes is $(c, x_c, y_c, w, h)$, where $c$ marks the category and $(x_c, y_c)$ denotes the positions of the bounding boxes' center in the image. $w$ and $h$ are the width and height of the bounding box, respectively. Some samples of annotated original images in SAD are shown in Fig. 4.

### C. Dataset Splits

To ensure that the distributions of the training and testing sets are approximately similar, we randomly select 420 images as the training set and the remaining 204 images as the testing set. It is worth mentioning that when we divide the dataset, different images of the same airport are uniformly split into training or testing sets. We will publicly provide all the original images with ground truth for the training set and testing set.

### D. Various Sizes of Airport Targets

We count the number of pixels in each airport bounding box, which we call pixel size for short, as a measurement for airport size. Fig. 2(a) illustrates the distribution of pixel sizes for instances in our dataset. We can see that the pixel sizes of airport targets are mostly in the range of 20,000 to 400,000 pixels. The smallest airport target contains 26,244 pixels and the largest has 114,250 pixels, which occupies 27.229% of the entire image size. We divide all the airports in SAD into three splits according to their number of pixels: small for range from 0 to 100,000, middle for range from 100,000 to 300,000, and large for range above 100,000. Table I illustrates the percentages of the three intervals in the SAD. The great variation in the pixel sizes of different airport targets makes it more challenging for the current detection methods.

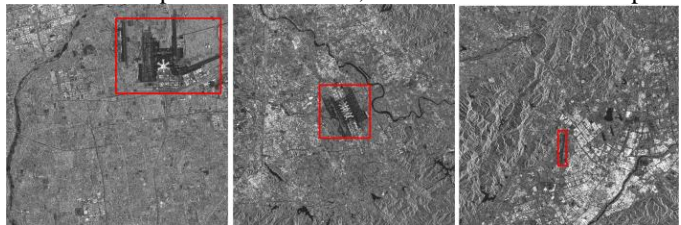

**Fig. 4.** Samples of annotated images in SAD.

TABLE I
AIRPORT AREA SIZE DISTRIBUTION IN SAD

| Intervals | Small | Middle | Large |
|---|---|---|---|
| Proportion | 52.4% | 39.7% | 7.9% |

### E. Various Aspect Ratio of Airports

Aspect ratio is an important factor affecting the detection accuracy of anchor-based object detection models, such as Faster RCNN, Yolo V3, etc. To facilitate SAD users to better set the aspect ratio of anchor, we count the aspect ratios of all airport targets in the SAD. Fig. 2(b) illustrates the distribution of aspect ratios for airport targets in our dataset. We can see that targets vary greatly in aspect ratio. The aspect ratios of the airport targets in the SAD are distributed between 0 and 5.2, with a relatively large cluster at around 0.8. Moreover, there are a large number of targets with a large aspect ratio in our dataset. Fig. 5 shows three airport targets with aspect ratios of the different. Finally, the wide distribution of aspect ratios enables the models to have high generalization performance.

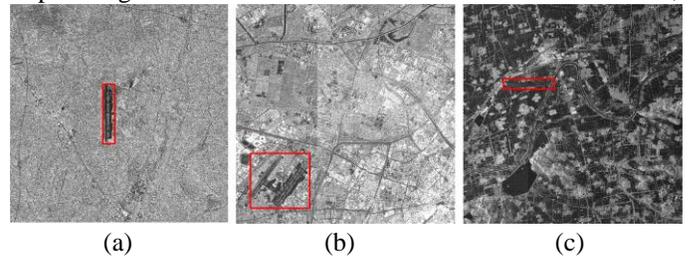

(a) (b) (c)

**Fig. 5.** Various aspect ratio of airports. (a)Aspect ratio=5.2. (b)Aspect ratio=0.8. (c)Aspect ratio=0.2.

### F. Various Azimuth of Airports

As shown in Table II, our dataset achieves a good balance in the airports of different directions., which is significantly helpful for learning a robust detector. Moreover, our dataset is closer to real scenes, because it is common to see airports in all kinds of orientations in the real world.

TABLE II
AIRPORT AZIMUTH INFORMATION STATISTICS

| Degree | 0°-45° | 45°-90° | 90°-135° | 135°-180° |
|---|---|---|---|---|
| Quantity | 133 | 143 | 180 | 168 |
| Proportion | 21.3% | 23.0% | 28.7% | 27.0% |

## V. CONCLUSION

In this paper, we first make a public SAR airport dataset. It provides a benchmark resource for developing airport detection algorithms or other relevant tasks.

ACKNOWLEDGMENT

The dataset is publicly available online: https://github.com/search?q=user%3ALongging+SAD.

REFERENCES

[1] "IEE Colloquium on 'Hough Transforms' (Digest No.106)," IEE Colloquium on Hough Transforms, 1993, pp. 0_1-.
[2] Von Gioi R G, Jakubowicz J, Morel J M, et al. LSD: a line segment detector[J]. Image Processing On Line, 2012, 2: 35-55.
[3] Wang W, Liu L, Hu C, et al. Airport detection in SAR image based on perceptual organization[C]//2011 International Workshop on Multi-Platform/Multi-Sensor Remote Sensing and Mapping. IEEE, 2011: 1-5.
[4] Xiong W, Zhong J, Zhou Y. Automatic recognition of airfield runways based on Radon transform and hypothesis testing in SAR images[C]//Proceedings of 2012 5th Global Symposium on Millimeter-Waves. IEEE, 2012: 462-465.
[5] Kou Z, Shi Z, Liu L. Airport detection based on line segment detector[C]//2012 International Conference on Computer Vision in Remote Sensing. IEEE, 2012: 72-77.
[6] Tang G, Xiao Z, Liu Q, et al. A novel airport detection method via line segment classification and texture classification[J]. IEEE Geoscience and Remote Sensing Letters, 2015, 12(12): 2408-2412.
[7] Budak Ü, Halıcı U, Şengür A, et al. Efficient airport detection using line segment detector and fisher vector representation[J]. IEEE Geoscience and Remote Sensing Letters, 2016, 13(8): 1079-1083.
[8] Zhu D, Wang B, Zhang L. Two-way saliency for airport detection in remote sensing images[C]//2014 International Conference on Audio, Language and Image Processing. IEEE, 2014: 526-531.
[9] Yao X, Han J, Guo L, et al. A coarse-to-fine model for airport detection from remote sensing images using target-oriented visual saliency and CRF[J]. Neurocomputing, 2015, 164: 162-172.
[10] Liu N, Cui Z, Cao Z, et al. Airport detection in large-scale SAR images via line segment grouping and saliency analysis[J]. IEEE Geoscience and Remote Sensing Letters, 2018, 15(3): 434-438.
[11] Li Z, Liu Z, Shi W. Semiautomatic airport runway extraction using a line-finder-aided level set evolution[J]. IEEE Journal of Selected Topics in Applied Earth Observations and Remote Sensing, 2014, 7(12): 4738-4749.
[12] Tao C, Tan Y, Cai H, et al. Airport detection from large IKONOS images using clustered SIFT keypoints and region information[J]. IEEE Geoscience and Remote Sensing Letters, 2010, 8(1): 128-132.
[13] Aytekin Ö, Zöngür U, Halici U. Texture-based airport runway detection[J]. IEEE Geoscience and Remote Sensing Letters, 2012, 10(3): 471-475.
[14] Yang J Y, Li H C, Hu W S, et al. Adaptive Cross-Attention-Driven Spatial-Spectral Graph Convolutional Network for Hyperspectral Image Classification[J]. IEEE geoscience and remote sensing letters, 2021.
[15] Yue Z, Gao F, Xiong Q, et al. A novel semi-supervised convolutional neural network method for synthetic aperture radar image recognition[J]. Cognitive Computation, 2021, 13(4): 795-806.
[16] Zhang S, Lin Y, Zhang X, et al. Airport automatic detection in large space-borne SAR imagery[J]. Journal of Systems Engineering and Electronics, 2010, 21(3): 390-396.
[17] Li H C, Hu W S, Li W, et al. A³CLNN: Spatial, spectral and multiscale attention ConvLSTM neural network for multisource remote sensing data classification[J]. IEEE Transactions on Neural Networks and Learning Systems, 2020.
[18] Wang X, Lv Q, Wang B, et al. Airport detection in remote sensing images: A method based on saliency map[J]. Cognitive neurodynamics, 2013, 7(2): 143-154.
[19] Zhu D, Wang B, Zhang L. Airport target detection in remote sensing images: A new method based on two-way saliency[J]. IEEE Geoscience and Remote Sensing Letters, 2015, 12(5): 1096-1100.
[20] Zhang L, Zhang Y. Airport detection and aircraft recognition based on two-layer saliency model in high spatial resolution remote-sensing images[J]. IEEE Journal of Selected Topics in Applied Earth Observations and Remote Sensing, 2016, 10(4): 1511-1524.
[21] F. Ma, F. Zhang, D. Xiang, Q. Yin and Y. Zhou, "Fast Task-Specific Region Merging for SAR Image Segmentation," in IEEE Transactions on Geoscience and Remote Sensing, vol. 60, pp. 1-16, 2022, Art no. 5222316, doi: 10.1109/TGRS.2022.3141125.
[22] F. Ma, F. Zhang, Q. Yin, D. Xiang and Y. Zhou, "Fast SAR Image Segmentation With Deep Task-Specific Superpixel Sampling and Soft Graph Convolution," in IEEE Transactions on Geoscience and Remote Sensing, vol. 60, pp. 1-16, 2022, Art no. 5214116, doi: 10.1109/TGRS.2021.3108585.
[23] Liu N, Cao Z, Cui Z, et al. Multi-layer abstraction saliency for airport detection in SAR images[J]. IEEE Transactions on Geoscience and Remote Sensing, 2019, 57(12): 9820-9831.
[24] Zhang P, Niu X, Dou Y, et al. Airport detection on optical satellite images using deep convolutional neural networks[J]. IEEE Geoscience and Remote Sensing Letters, 2017, 14(8): 1183-1187.
[25] Xu Y, Zhu M, Li S, et al. End-to-end airport detection in remote sensing images combining cascade region proposal networks and multi-threshold detection networks[J]. Remote Sensing, 2018, 10(10): 1516.
[26] Li S, Xu Y, Zhu M, et al. Remote sensing airport detection based on end-to-end deep transferable convolutional neural networks[J]. IEEE Geoscience and Remote Sensing Letters, 2019, 16(10): 1640-1644.
[27] Sun X, Wang P, Yan Z, et al. FAIR1M: A benchmark dataset for fine-grained object recognition in high-resolution remote sensing imagery[J]. ISPRS Journal of Photogrammetry and Remote Sensing, 2022, 184: 116-130.
[28] Cai B, Jiang Z, Zhang H, et al. Airport detection using end-to-end convolutional neural network with hard example mining[J]. Remote Sensing, 2017, 9(11): 1198.
[29] Yin S, Li H, Teng L. Airport detection based on improved faster RCNN in large scale remote sensing images[J]. Sensing and Imaging, 2020, 21(1): 1-13.
[30] Chen F, Ren R, Van de Voorde T, et al. Fast automatic airport detection in remote sensing images using convolutional neural networks[J]. Remote Sensing, 2018, 10(3): 443.
[31] Zeng F, Cheng L, Li N, et al. A hierarchical airport detection method using spatial analysis and deep learning[J]. Remote Sensing, 2019, 11(19): 2204.
[32] Chen L, Tan S, Pan Z, et al. A new framework for automatic airports extraction from SAR images using multi-level dual attention mechanism[J]. Remote Sensing, 2020, 12(3): 560.
[33] Tan S, Chen L, Pan Z, et al. Geospatial Contextual Attention Mechanism for Automatic and Fast Airport Detection in SAR Imagery[J]. IEEE Access, 2020, 8: 173627-173640.
[34] Zaidi S S A, Ansari M S, Aslam A, et al. A Survey of Modern Deep Learning based Object Detection Models[J]. arXiv preprint arXiv:2104.11892, 2021.
[35] Girshick R, Donahue J, Darrell T, et al. Rich feature hierarchies for accurate object detection and semantic segmentation[C]//Proceedings of the IEEE conference on computer vision and pattern recognition. 2014: 580-587.
[36] Girshick R. Fast r-cnn[C]//Proceedings of the IEEE international conference on computer vision. 2015: 1440-1448.
[37] Ren S, He K, Girshick R, et al. Faster r-cnn: Towards real-time object detection with region proposal networks[J]. Advances in neural information processing systems, 2015, 28: 91-99.
[38] He K, Gkioxari G, Dollár P, et al. Mask r-cnn[C]//Proceedings of the IEEE international conference on computer vision. 2017: 2961-2969.
[39] Redmon J, Farhadi A. Yolov3: An incremental improvement[J]. arXiv preprint arXiv:1804.02767, 2018.
[40] Bochkovskiy A, Wang C Y, Liao H Y M. Yolov4: Optimal speed and accuracy of object detection[J]. arXiv preprint arXiv:2004.10934, 2020.